\pdfoutput=1

\documentclass[11pt]{article}

\usepackage[]{acl}

\usepackage{acl}

\usepackage{times}
\usepackage{latexsym}
\usepackage{amssymb}
\usepackage{xcolor}
\usepackage{color,colortbl}
\usepackage{amsmath}
\usepackage{multirow}
\usepackage{graphicx}
\usepackage{paralist}
\usepackage{booktabs}
\usepackage{array}
\usepackage{multicol}
\usepackage{subcaption}
\usepackage{enumitem}
\usepackage{tabularx}
\usepackage{hyperref}

\usepackage{algorithm,algpseudocode}
\usepackage{newfloat}
\usepackage{listings}

\definecolor{Gray}{gray}{0.9}

\usepackage[T1]{fontenc}

\usepackage[utf8]{inputenc}

\usepackage{microtype}

\title{{P}ost-{H}oc {A}nswer {A}ttribution for {G}rounded and {T}rustworthy {L}ong {D}ocument {C}omprehension: {T}ask, {I}nsights, and {C}hallenges}

\author {
    Abhilasha Sancheti\textsuperscript{$\dagger$}\thanks{This work was done when the author was at Adobe.}, Koustava Goswami\textsuperscript{$\ddagger$}, Balaji Vasan Srinivasan\textsuperscript{$\ddagger$} \\
    \textsuperscript{$\dagger$}University of Maryland, College Park 
    \textsuperscript{$\ddagger$}Adobe Research\\
   \small{\href{mailto:sancheti@umd.edu}{\tt \textcolor{black}{sancheti@umd.edu}}}, 
    \small{\tt \{koustavag,balsirini\}@adobe.com}
}

\begin{document}

\maketitle

\begin{abstract}
Attributing answer text to its source document for information-seeking questions is crucial for building trustworthy, reliable, and accountable systems. We formulate a new task of post-hoc answer attribution for long document comprehension (LDC). Owing to the lack of long-form abstractive and information-seeking LDC datasets, we refactor existing datasets to assess the strengths and weaknesses of existing retrieval-based and proposed answer decomposition and textual entailment-based optimal selection attribution systems for this task. We throw light on the limitations of existing datasets and the need for datasets to assess the actual performance of systems on this task.
\end{abstract}
\section{Introduction}
Users now benefit from the help of automatic question-answering (QA) systems on a day-to-day basis when faced with an information need.
Such systems are integrated into search engines (\textit{e.g.}, BingAI\footnote{\url{https://www.microsoft.com/en-gb/bing?form=MW00X7}}) and digital assistants (\textit{e.g.}, ChatGPT). 
However, such systems are prone to generating answers lacking sufficient grounding to knowledge sources~\citep{dziri-etal-2022-origin,ji2023survey}, leading to the risks of misinformation and hallucination~\citep{metzler2021rethinking,shah2022situating,huo2023retrieving}. Therefore, attributing the generated answers to the respective sources is crucial for building trustworthy, reliable, verifiable, and accountable systems~\citep{bohnet2022attributed,huang2023citation,rashkin2023measuring,yue2023automatic}; by allowing users to verify outputs.

\begin{table}[t!]
\centering
\scriptsize
\begin{tabular}{p{0.93\columnwidth}}
\toprule
\rowcolor{Gray}
\multicolumn{1}{c}{\textbf{Input}}\\
\textbf{Question:} When does the next assasins creed come out?\\
\textbf{Document:} {[1]} Ubisoft has announced that its next Assassin's Creed game will be revealed in September 2022. \\
{[2]} Ubisoft shared the first trailer for the game on Saturday.\\
{[3]} Assassin's Creed Mirage, the next entry in Ubisoft's long-running action-adventure series, will arrive in 2023.\\
{[4]} The publisher announced the release date today during its Ubisoft Forward event. \dots\\
\textbf{Answer:} The next Assassin's Creed game, Assassin's Creed Mirage, will arrive in 2023 according to Ubisoft's announcement during its Ubisoft Forward event. It will be released for Xbox \dots The game will be revealed in September 2022.\\
\midrule
\rowcolor{Gray}
\multicolumn{1}{c}{\textbf{Output}}\\
\textbf{Attributed answer:} The next Assassin's Creed game, Assassin's Creed Mirage, \dots Ubisoft's announcement during its Ubisoft Forward event {[3,4]} \dots The game will be revealed in September 2022 {[1]}.\\
\bottomrule
\end{tabular}
\caption{An example taken from reformulated verifiability  dataset~\citep{liu2023evaluating} that
includes a question, a document,\protect\footnotemark~and an answer as inputs, and the document-grounded attributions for each sentence (some may not have any attribution) in the answer as output. } \label{tab:task-input-output}
\end{table}
\footnotetext{A subset of sentences is shown due to space constraints.}
Existing works mainly consider generating attributed text in open-ended settings. These attributions are generated along with the answers either one per answer paragraph~\citep{bohnet2022attributed,hu2024benchmarking} or per answer sentence~\citep{gao-etal-2023-rarr,gao2023enabling,malaviya2023expertqa}. Evidence retrieval is used to select an answer in reading comprehension setting~\citep{wang2019evidence,yadav-etal-2020-unsupervised,cui2022expmrc} for short and extractive answers.
Attribution becomes challenging when answers are abstractive such that each sentence could be composed of multiple sentences in the source document, requiring more sophisticated approaches. To address this gap, we aim to identify fine-grained attributions (\textit{i.e.}, sentences grounded in a provided long document) for each sentence (unlike paragraph or article) of a long-form abstractive answer to an information-seeking question asked over a user-provided document (closed-domain). Such fine-grained attributions can lead to more trustworthy, reliable, and accountable systems.
Specifically, we propose a new task (Table~\ref{tab:task-input-output}) of \textbf{post-hoc answer attribution for long document comprehension} wherein the input to a system is a \textbf{(question, answer, document)} triplet, and output is an \textbf{attributed answer} consisting of pointers to sentences in the document that provide supporting evidence for each sentence in the answer. 

Building systems for this task is challenging due to the unavailability of appropriate datasets as answers in existing information-seeking reading comprehension datasets (\textit{e.g.}, \citealp{dasigi2021dataset}) are short and extractive. Moreover, obtaining attribution annotations is cognitively demanding, labor-intensive, and expensive as it requires expertise~\citep{kamalloo2023hagrid}. Thus, we \begin{inparaenum}[(a)]
\item{propose to reformulate existing datasets curated for evaluating citation verifiability in generative search engines~\citep{liu2023evaluating}, and generating attributed explanations in generative information-seeking systems~\citep{kamalloo2023hagrid}, and}
\item{assess the feasibility of using existing textual entailment models by proposing \texttt{ADiOSAA}-- consisting of an answer decomposer and a textual entailment-based attributor that uses an optimal selection strategy to find attributions for each sentence of an answer.}
\end{inparaenum}

This work \textbf{contributes} the following: \begin{inparaenum}[(1)]
\item{introduces the task of \textbf{post-hoc answer attribution for LDC} for building trustworthy, verifiable, reliable, and accountable QA systems~(\textsection{\ref{subsec:dataset}});}
\item{reformulates existing datasets for this task, owing to the lack of availability of long-form abstractive reading comprehension datasets (\textsection{\ref{subsec:dataset}}), and}
\item{assesses the strengths and weaknesses of existing retrieval-based systems, and proposed answer decomposition and textual entailment-based optimal selection system, \texttt{ADiOSAA}  (\textsection{\ref{sec:method}}), by adopting information retrieval measures (\textsection{\ref{sec:evaluation}}).}
\end{inparaenum}

\section{Adapting existing datasets for our task} \label{subsec:dataset} 
\paragraph{Task Definition}We formalize the task of post-hoc answer attribution for long document comprehension as: given a query $Q$, a set of sentences $S=s_1, \dots, s_n$ from document $D$ (namely, source sentences), and an answer (either generated from a system or ground-truth) to query $Q$, the goal is to identify supporting sentences (namely, attributions) $s_i\in S$ for each answer sentence $a_i\in A=a_1, \dots, a_m$ (may be attributed to multiple source sentences or none).
Since there are no datasets that match the needs of our task, we propose to reformulate the Citation Verifiability dataset \cite{liu2023evaluating} and Hagrid dataset \cite{kamalloo2023hagrid} for the proposed task. 
\paragraph{Reformulation Citation Verifiability Dataset}
 Citation verifiability dataset~\citep{liu2023evaluating} consists of questions from NaturalQuestions~\citep{kwiatkowski-etal-2019-natural} and ELI5~\citep{fan-etal-2019-eli5} and answers are generated from different generative search engines; Bing Chat, NeevaAI, perplexity.ai, and YouChat. These answers are embedded with inline citations pointing to the web pages. Human annotators were shown a question and a verification-worthy sentence from the generated answer with its corresponding generated citations and were asked to judge if the citations \textit{fully, partially, or do not support} the sentence. For sentences that are \textit{fully} supported, annotators also provide sentences on the webpage that support the answer sentence. In this open-domian setup, the citations in an answer may belong to multiple web pages.
 To obtain a pseudo document for a question, we focus on questions anchored to a given document by combining fully supported web page contents cited for sentences.
 Hence, we have a corpus with questions, answers, a document to which questions are grounded, and ground truth attributions for sentences in an answer.
\begin{table}[t]
\centering
\scriptsize
\resizebox{0.99\columnwidth}{!}{
\begin{tabular}{l|ccccc}
\toprule
\textbf{Split} &\multicolumn{1}{m{0.7cm}}{\textbf{Size}} & \multicolumn{1}{p{1cm}}{\textbf{Avg. No. of source sentences}} & \multicolumn{1}{p{1.1cm}}{\textbf{Avg. No. of attributions per sentence}} & \multicolumn{1}{p{1.1cm}}{\textbf{Avg. No. of sentences per answer}} & \multicolumn{1}{p{0.98cm}}{\textbf{Avg. No. of answers per question}}\\
\midrule
\multicolumn{6}{c}{\textbf{Verifiability/Hagrid}} \\
\midrule
Train &$1138/1922$&$128.58/2.82$&$1.45/1.26$&$2.11/1.63$&$2.63/1.67$\\
Dev & $146/716$&$141.68/2.83$&$1.49/1.40$&$2.18/1.71$&$2.56/1.84$\\
Test & $136/-$&$130.03/-$&$1.60/-$&$2.13/-$&$2.75/-$\\
\bottomrule
\end{tabular}
}
\caption{Dataset statistics. No test set in Hagrid.} \label{tab:data-stats}
\end{table}
\paragraph{Reformulating Hagrid Dataset}
\citet{kamalloo2023hagrid} introduced 
Hagrid which is constructed based on human and LLM collaboration by first automatically collecting attributed answers (for information-seeking questions in MIRACL \citep{zhang2022making} dataset) that follow an inline citation style using GPT-$3.5$. Then, asking human annotators to evaluate the LLM answers based on informativeness and attributability. We establish benchmarks for this dataset by considering the LLM-generated answers to be the gold-answers required as input (as opposed to the task formulation of Hagrid, wherein output is an attributed answer), attributability annotations as attributions for sentences in an answer, and labeled relevant passages as the document. We provide dataset statistics in Table~\ref{tab:data-stats}.

\section{\textbf{A}nswer \textbf{D}ecompos\textbf{i}tion and \textbf{O}ptimal \textbf{S}election for \textbf{A}nswer \textbf{A}ttribution} \label{sec:method}

\begin{figure}[t!]
    \centering
    \includegraphics[width=\columnwidth]{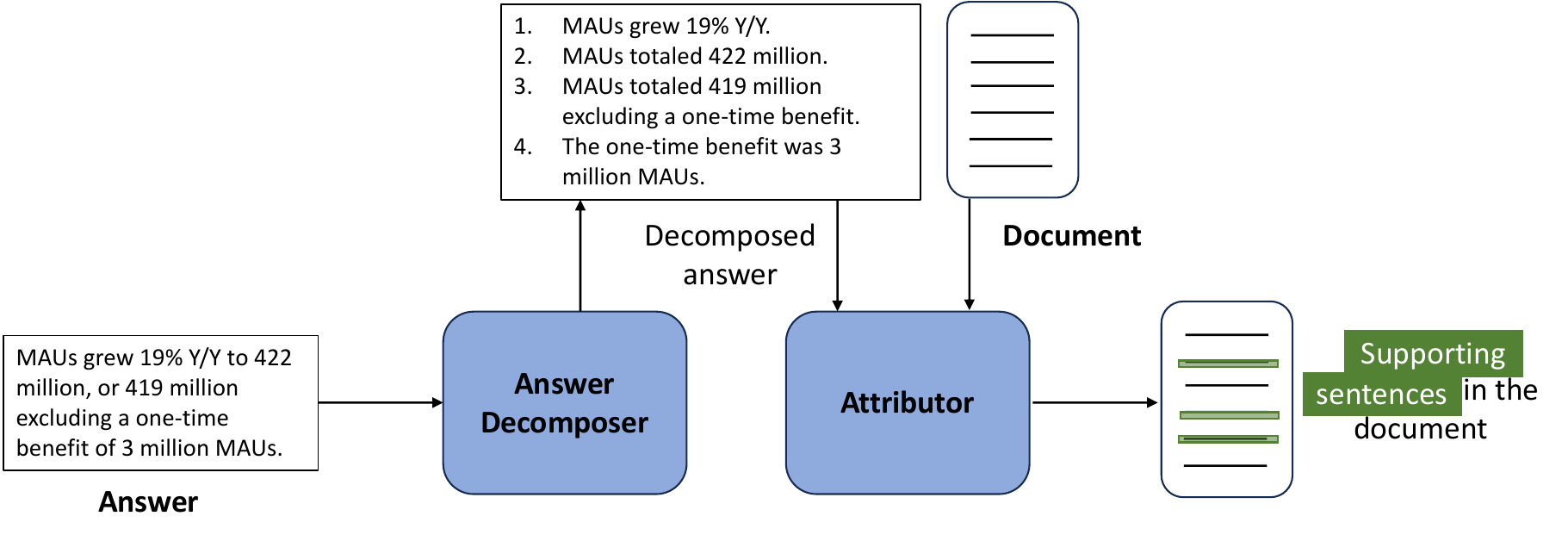}
    \caption{Overview of proposed answer attribution system, \texttt{ADiOSAA}. The \textbf{answer decomposer} breaks the given answer into \textit{information units}, and the \textbf{attributor} finds the supporting sentences as attributions for each \textit{information unit} in the answer.}
    \label{fig:proposed}
\end{figure}
We propose an \textbf{A}nswer \textbf{D}ecompos\textbf{i}tion and \textbf{O}ptimal \textbf{S}election \textbf{A}nswer \textbf{A}ttribution system for the introduced task. \texttt{ADiOSAA} consists of two components (Figure~\ref{fig:proposed}):
\begin{inparaenum}[(1)]
\item{An \textbf{answer decomposer} to break each sentence of an answer into one or more \textit{information units}~\citep{nenkova-passonneau-2004-evaluating,stanovsky-etal-2018-supervised,ernst-etal-2021-summary} as we believe that an answer sentence is composed of information from multiple sentences in the input document.}
\item{An \textbf{attributor} to find supporting sentences in the document for a given \textit{information unit} in the answer sentence.}
\end{inparaenum}

\paragraph{Answer Decomposer} We prompt (``Please breakdown the following sentence into independent facts: ..") ChatGPT~\citep{chatgpt} to decompose the given answer into its information units, following \citet{min2023factscore} who found such decompositions to be effective and close to human. This decomposition resembles past frameworks derived from OpenIE~\citep{stanovsky-etal-2018-supervised,ernst-etal-2021-summary} or
Pyramid~\citep{nenkova-passonneau-2004-evaluating,shapira-etal-2019-crowdsourcing}, but avoids relying on annotated data and achieves greater flexibility by using ChatGPT. Such decomposition to information units has been successfully used for claim-verification~\citep{kamoi2023wice} and propositional semantic representations~\citep{chen2023sub}.

\paragraph{Attributor} Once the answer is decomposed into its information units, each unit needs to be mapped to sentences in the input document to provide the desired attributions. We pose this task of finding supporting sentences in the document for a given information unit as a textual entailment task. Textual entailment is the task of identifying if a given premise ($\mathcal{P}$) entails or does not entail the given hypothesis ($\mathcal{H}$). For our purpose, we consider sentence(s) in the document as the premise and an information unit as the hypothesis. We use RoBERTa-L~\citep{liu2019roberta} pretrained\footnote{We use the official code and trained model available at \url{https://github.com/salesforce/DocNLI}.} on DocNLI~\citep{yin-etal-2021-docnli} dataset (contains paragraph-level (premise, hypothesis) pairs, see \textsection\ref{sec:app-docnli} for more details) as the entailment model (attributor) to predict if the given information unit can be inferred from the given sentence(s) from the document.  
\begin{algorithm}[t!]
\small
  \caption{Optimal Selection Algorithm}
  \label{alg:optimal}
  \begin{algorithmic}[1]
    \State \textbf{Inputs:} Information unit (\textit{iu}), $D={d_1,d_2\ldots d_n}$, $Attr(\mathcal{P},\mathcal{H})$, $\delta$ 
     \State \textbf{Outputs:} $L$ = A list of supporting sentences in $D$ which together attribute \textit{iu}
    \State ${L} \gets [ ]$, RS $\gets D$, prev\_score $\gets -1$  
    \hfill // \text{RS: remaining sentences}; Initialization
    \While{RS is not empty}
    \State $\text{curr\_score} \gets \operatorname*{\max}_{d_i \in \text{RS} } \text{Attr(}L+ d_i, \textit{iu})$
    \State $d_{max} \gets \arg\max_{d_i \in \text{RS} }\text{Attr}(L + d_i, \textit{iu})$
             \If{curr\_score > prev\_score + $\delta$}
                    \State $L \mathrel{+}= d_{max}$
                    \State RS $\mathrel{-}=d_{max}$ 
                    \State prev\_score = curr\_score
                   
            \Else 
                    \State break
                \EndIf
\EndWhile
  \end{algorithmic}
\end{algorithm}
\begin{table*}[h]
\centering
\scriptsize
\resizebox{1.999\columnwidth}{!}{
\begin{tabular}{l|ccc|ccc}
\toprule
\multirow{2}{*}{\centering{\textbf{Model}}} & \multicolumn{3}{c|}{\centering{\textbf{Verifiability}}} & \multicolumn{3}{c}{\centering{\textbf{Hagrid}}}\\
\cline{2-4} \cline{5-7}\\
& \textbf{(P/R/F1)@1}& \textbf{(P/R/F1)@2} & \textbf{(P/R/F1)@4} & \textbf{(P/R/F1)@1}& \textbf{(P/R/F1)@2} & \textbf{(P/R/F1)@4}\\
\midrule
BM25&	$0.669/0.529/0.567$	&	$0.443/0.648/0.499$	&	$0.270/0.722/0.369$	&	$0.815/0.686/0.722$	&	$0.740/0.919/0.788$	&	$0.678/0.990/0.760$	\\
GTR&	$0.656/0.511/0.550$	&	$0.432/0.623/0.483$	&	$0.270/0.723/0.371$	&	$0.899/0.768/0.804$	&	$0.744/0.918/0.790$	&	$0.677/0.987/0.759$	\\
MonoT5&	$\mathbf{0.698}/\mathbf{0.552}/\mathbf{0.593}$ &		$0.466/\mathbf{0.675}/\mathbf{0.522}$	&	$0.284/\mathbf{0.757}/0.389$	&	$\mathbf{0.962}/\mathbf{0.827}/\mathbf{0.864}$	&	$0.763/\mathbf{0.946}/\mathbf{0.811}$	&	$0.680/\mathbf{0.993}/0.762$	\\
\texttt{ADiOSAA}&	$0.545/0.428/0.459$	&	$\mathbf{0.484}/0.546/0.487$	&	$\mathbf{0.476}/0.604/\mathbf{0.499}$	&	$0.856/0.734/0.768$	&	$0.848/0.810/0.799$	&	$0.848/0.817/\mathbf{0.801}$	\\
\texttt{ADiOSAA} - D&	$0.473/0.388/0.412$	&	$0.445/0.418/0.412$	&	$0.442/0.418/0.411$	&	$0.869/0.749/0.782$	&	$\mathbf{0.861}/0.758/0.783$	&	$\mathbf{0.861}/0.758/0.783$	\\
\texttt{ADiOSAA} - OS&	$0.375/0.295/0.317$	&	$0.280/0.333/0.284$	&	$0.256/0.360/0.276$	&	$0.793/0.679/0.710$	&	$0.745/0.783/0.736$	&	$0.743/0.830/0.752$	\\
\texttt{ADiOSAA} - D - OS&	$0.269/0.234/0.243$	&	$0.269/0.234/0.243$		&	$0.269/0.234/0.243$		&	$0.567/0.466/0.494$	&	$0.567/0.466/0.494$	&	$0.567/0.466/0.494$	\\
\bottomrule
\end{tabular}
}
\caption{Evaluation results: \texttt{ADiOSAA} systems use top $150$ source sentences (see Table~\ref{tab:add-results-verifiability} in Appendix for results with GTR, MonoT5, and all the source sentences) retrieved using BM25 for the Verifiability dataset. D denotes Answer Decomposer, and OS refers to Optimal Selection.} \label{tab:results}
\end{table*}
\paragraph{Optimal Selection} An answer sentence could be attributed to multiple sentences in the provided document when: \begin{inparaenum}[(a)]
 \item{the same information is available in the document at multiple places, and}
 \item{pieces of information in the answer sentence is available in different parts of the document.}
\end{inparaenum}
(a) can be solved by considering the top $k$ (premise hypothesis) pairs where the premise is the sentence from the document and the hypothesis is the sentence or information unit of the answer. To solve (b), it is required to check if a sentence or information unit of an answer can be entailed from a combination of sentences in the document as a premise. However, this becomes computationally expensive; for a document consisting of $N$ sentences, there will be $2^N$ combinations. To address this issue, we propose an optimal selection approach that greedily selects sentences from the document that has the maximum probability of entailment as described in Algorithm~\ref{alg:optimal}. Attr($\mathcal{P}$,$\mathcal{H}$) refers to DocNLI-based attributor which takes sentences from the input document and the information unit (or sentence in an answer) and outputs the probability of entailment of $\mathcal{H}$ from $\mathcal{P}$. For each information unit in a sentence, Algorithm~\ref{alg:optimal} iteratively selects a sentence from the set of remaining source sentences that maximizes the probability of entailment until the entailment score keeps increasing above a threshold $\delta$ as compared to that in the previous iteration.

 We reorder the attributions for each information unit based on their score and deduplicate (as different information units may be attributed to the same source sentence) them to obtain the predicted attributions for each sentence of an answer.

\section{Evaluation} \label{sec:evaluation}
As answer sentence attribution to sentences in the source document could also be considered as an information retrieval task, we benchmark the performance of a range of retrieval-based systems: \textbf{(1) BM25 (sparse)}, \textbf{(2) GTR (dense)}, and \textbf{(3) MonoT5}, considering an answer sentence as the query, and the sentences/passages from the input document as the document (refer to \textsection{\ref{sec:app-baselines}}). Because our task assumes the answer as an input, inline attribution-based systems like vanilla LLM prompting~\cite{tay2022transformer,weller2023according} and retrieve-and-read-based systems~\citep{guu2020retrieval,borgeaud2022improving,izacard2022few} do not fit here.
For the Verifiability dataset, \texttt{ADiOSAA} system and its variants use top $150$ retrieved sentences as the source sentences. As Hagrid has only $2.83$ passages per question in total, we consider all the passages as the source sentences. 
Additionally, we perform ablation experiments to demonstrate the importance of decomposition and optimal selection in \texttt{ADiOSAA} in the following ways.
\paragraph{\texttt{ADiOSAA} - D} considers an answer sentence as the information unit instead of decomposing it. This system establishes the importance of the answer decomposer in \texttt{ADiOSAA}.
\paragraph{\texttt{ADiOSAA} - OS} decomposes each answer sentence into its information units, and then ranks source sentences based on their entailment probabilities from the Attr($\mathcal{P}$,$\mathcal{H}$) for each information unit. To obtain attributions for each sentence of the answer, it deduplicates and reorders the attributions for all the information units of the sentence based on the entailment probabilities.
\paragraph{\texttt{ADiOSAA} - D - OS} neither uses the answer decomposer or the optimal selection algorithm rather for each sentence in the answer, it ranks source sentences based on their entailment probabilities from the Attr($\mathcal{P}$,$\mathcal{H}$). This system demonstrates the effectiveness of both the components in \texttt{ADiOSSA}.

\paragraph{Evaluation Measures} We report precision (P), recall (R), and F1@k $\in \{1,2,4\}$ predicted attributions per sentence of an answer\footnote{We filter out the instances where answer sentences were extracted directly from the documents.} for the test set of Verifiability dataset and development set of the Hagrid dataset (as no test set is available). We tune the threshold for attributor's entailment probability (=$0.5$) and $\delta$ (=$0.3$) in Algorithm~\ref{alg:optimal} based on the Verifiability development set. 

\section{Results and Discussion} \label{sec:results}
While MonoT5-based retrieval system outperforms (Table~\ref{tab:results}) others for the top-1 prediction, \texttt{ADiOSAA} variants attain the highest precision when top 2 or 4 predictions are considered. Having a high precision for top 2 or 4 predictions is important as the mean number of attributions per sentence $>$ 1 (see Table~\ref{tab:data-stats}) and with the increase in the number of predictions, recall may increase or remain the same however, precision may increase, decrease, or stay the same.
\texttt{ADiOSAA} variants retain higher precision (as compared to retrieval-based systems) even with the increase in the number of predictions, indicating that retrieval-based systems are good at retrieving one attribution correctly but fail for the second (or more) one compared to our systems. This shows that our systems capture abstractive and compositional attributions more correctly. 
Optimal selection results in a significant improvement. Higher gains due to optimal selection under no decomposition (difference between \texttt{ADiOSAA}-D and \texttt{ADiOSAA}-D-OS) than under decomposition (difference between \texttt{ADiOSAA} and \texttt{ADiOSAA}-OS) shows that the answer sentence is composed of multiple document sentences which are better captured with optimal selection. However, under decomposition, it is more likely that now the decomposed units could be attributed to a single sentence in the document. Decomposition also helps in better predictions (compare \texttt{ADiOSAA}-OS with \texttt{ADiOSAA}-D-OS) showing that compositional answers have multiple attributions to different sentences in the input document. 
Further, due to a small number of source sentences (avg. $2.83$) in Hagrid, the precision and recall values are higher as compared to that in the Verifiability dataset. 

Good performance of retrieval-based systems indicate that the existing datasets are less abstractive for long-form comprehension, suggesting the need for research in creating more challenging datasets to foster the development of trustworthy, reliable, and accountable systems that can be used in real-world information-seeking scenarios.

\paragraph{Quality of Decompositions} Prior works have used ChatGPT for decomposing facts~\citep{min2023factscore} or claims~\citep{kamoi2023wice} and have shown it to perform reasonably well. We manually examine a subset of decompositions and find that the decomposer might sometimes over-decompose a simple sentence, or generate hallucinated information units (see Table~\ref{tab:decompose-examples} in the appendix for examples). We leave a careful analysis of error categories, and ways to mitigate hallucinations and over-decompositions for future work.
\section{Conclusion}
We introduce a task of post-hoc answer attribution for long document
comprehension, reformulate existing datasets, and asses the feasibility of existing textual entailment and retrieval-based systems in performing this task. 
Evaluation shows that retrieval-based systems are good at top one prediction however, our proposed answer decomposition and textual entailment-based optimal selection system, \texttt{ADiOSAA}, performs better when more than one predictions are considered. This further indicates the need for highly abstractive long-form reading comprehension datasets that can foster the development and evaluation of more sophisticated attribution systems.
\section{Limitations}
We note the following limitations of our work.
\begin{inparaenum}[(1)]
\item{The decompositions are obtained without taking into consideration the source document which might result in unnecessary answer decompositions. This issue can be resolved if the information units are explicitly constrained in the input document, and}
\item{\texttt{ADiOSAA} is a post-hoc inference time attribution system which uses off-the-shelf trained model, DocNLI. However, future work may consider developing supervised systems for performing the task on the verifiability dataset, and building end-to-end systems where decomposition and optimal selection may happen in an interactive manner.}
\item{We acknowledge the performance dependence of \texttt{ADiOSAA} on the Attributor. Further investigation into the impact of NLI model's performance on the final results is an avenue for future work.}
\end{inparaenum}

\bibliography{bibs/anthology,bibs/custom}

\begin{thebibliography}{42}
\expandafter\ifx\csname natexlab\endcsname\relax\def\natexlab#1{#1}\fi

\bibitem[{Bohnet et~al.(2022)Bohnet, Tran, Verga, Aharoni, Andor, Soares, Eisenstein, Ganchev, Herzig, Hui et~al.}]{bohnet2022attributed}
Bernd Bohnet, Vinh~Q Tran, Pat Verga, Roee Aharoni, Daniel Andor, Livio~Baldini Soares, Jacob Eisenstein, Kuzman Ganchev, Jonathan Herzig, Kai Hui, et~al. 2022.
\newblock Attributed question answering: Evaluation and modeling for attributed large language models.
\newblock \emph{arXiv preprint arXiv:2212.08037}.

\bibitem[{Borgeaud et~al.(2022)Borgeaud, Mensch, Hoffmann, Cai, Rutherford, Millican, Van Den~Driessche, Lespiau, Damoc, Clark et~al.}]{borgeaud2022improving}
Sebastian Borgeaud, Arthur Mensch, Jordan Hoffmann, Trevor Cai, Eliza Rutherford, Katie Millican, George~Bm Van Den~Driessche, Jean-Baptiste Lespiau, Bogdan Damoc, Aidan Clark, et~al. 2022.
\newblock Improving language models by retrieving from trillions of tokens.
\newblock In \emph{International conference on machine learning}, pages 2206--2240. PMLR.

\bibitem[{Chen et~al.(2023)Chen, Zhang, Chen, Zhou, Yu, Yu, Peng, Wang, Roth, and Yu}]{chen2023sub}
Sihao Chen, Hongming Zhang, Tong Chen, Ben Zhou, Wenhao Yu, Dian Yu, Baolin Peng, Hongwei Wang, Dan Roth, and Dong Yu. 2023.
\newblock Sub-sentence encoder: Contrastive learning of propositional semantic representations.
\newblock \emph{arXiv preprint arXiv:2311.04335}.

\bibitem[{Cui et~al.(2022)Cui, Liu, Che, Chen, and Wang}]{cui2022expmrc}
Yiming Cui, Ting Liu, Wanxiang Che, Zhigang Chen, and Shijin Wang. 2022.
\newblock Expmrc: explainability evaluation for machine reading comprehension.
\newblock \emph{Heliyon}, 8(4).

\bibitem[{Curation(2020)}]{curation2020}
Curation. 2020.
\newblock Curation. 2020. curation corpus base.

\bibitem[{Dasigi et~al.(2021)Dasigi, Lo, Beltagy, Cohan, Smith, and Gardner}]{dasigi2021dataset}
Pradeep Dasigi, Kyle Lo, Iz~Beltagy, Arman Cohan, Noah~A Smith, and Matt Gardner. 2021.
\newblock A dataset of information-seeking questions and answers anchored in research papers.
\newblock \emph{arXiv preprint arXiv:2105.03011}.

\bibitem[{Dziri et~al.(2022)Dziri, Milton, Yu, Zaiane, and Reddy}]{dziri-etal-2022-origin}
Nouha Dziri, Sivan Milton, Mo~Yu, Osmar Zaiane, and Siva Reddy. 2022.
\newblock \href {https://doi.org/10.18653/v1/2022.naacl-main.387} {On the origin of hallucinations in conversational models: Is it the datasets or the models?}
\newblock In \emph{Proceedings of the 2022 Conference of the North American Chapter of the Association for Computational Linguistics: Human Language Technologies}, pages 5271--5285, Seattle, United States. Association for Computational Linguistics.

\bibitem[{Ernst et~al.(2021)Ernst, Shapira, Pasunuru, Lepioshkin, Goldberger, Bansal, and Dagan}]{ernst-etal-2021-summary}
Ori Ernst, Ori Shapira, Ramakanth Pasunuru, Michael Lepioshkin, Jacob Goldberger, Mohit Bansal, and Ido Dagan. 2021.
\newblock \href {https://doi.org/10.18653/v1/2021.conll-1.25} {Summary-source proposition-level alignment: Task, datasets and supervised baseline}.
\newblock In \emph{Proceedings of the 25th Conference on Computational Natural Language Learning}, pages 310--322, Online. Association for Computational Linguistics.

\bibitem[{Fan et~al.(2019)Fan, Jernite, Perez, Grangier, Weston, and Auli}]{fan-etal-2019-eli5}
Angela Fan, Yacine Jernite, Ethan Perez, David Grangier, Jason Weston, and Michael Auli. 2019.
\newblock \href {https://doi.org/10.18653/v1/P19-1346} {{ELI}5: Long form question answering}.
\newblock In \emph{Proceedings of the 57th Annual Meeting of the Association for Computational Linguistics}, pages 3558--3567, Florence, Italy. Association for Computational Linguistics.

\bibitem[{Gao et~al.(2023{\natexlab{a}})Gao, Dai, Pasupat, Chen, Chaganty, Fan, Zhao, Lao, Lee, Juan, and Guu}]{gao-etal-2023-rarr}
Luyu Gao, Zhuyun Dai, Panupong Pasupat, Anthony Chen, Arun~Tejasvi Chaganty, Yicheng Fan, Vincent Zhao, Ni~Lao, Hongrae Lee, Da-Cheng Juan, and Kelvin Guu. 2023{\natexlab{a}}.
\newblock \href {https://doi.org/10.18653/v1/2023.acl-long.910} {{RARR}: Researching and revising what language models say, using language models}.
\newblock In \emph{Proceedings of the 61st Annual Meeting of the Association for Computational Linguistics (Volume 1: Long Papers)}, pages 16477--16508, Toronto, Canada. Association for Computational Linguistics.

\bibitem[{Gao et~al.(2023{\natexlab{b}})Gao, Yen, Yu, and Chen}]{gao2023enabling}
Tianyu Gao, Howard Yen, Jiatong Yu, and Danqi Chen. 2023{\natexlab{b}}.
\newblock Enabling large language models to generate text with citations.
\newblock \emph{arXiv preprint arXiv:2305.14627}.

\bibitem[{Guu et~al.(2020)Guu, Lee, Tung, Pasupat, and Chang}]{guu2020retrieval}
Kelvin Guu, Kenton Lee, Zora Tung, Panupong Pasupat, and Mingwei Chang. 2020.
\newblock Retrieval augmented language model pre-training.
\newblock In \emph{International conference on machine learning}, pages 3929--3938. PMLR.

\bibitem[{Hu et~al.(2024)Hu, Chen, Wu, Qi, Bi, Wu, and Pan}]{hu2024benchmarking}
Nan Hu, Jiaoyan Chen, Yike Wu, Guilin Qi, Sheng Bi, Tongtong Wu, and Jeff~Z Pan. 2024.
\newblock Benchmarking large language models in complex question answering attribution using knowledge graphs.
\newblock \emph{arXiv preprint arXiv:2401.14640}.

\bibitem[{Huang and Chang(2023)}]{huang2023citation}
Jie Huang and Kevin Chen-Chuan Chang. 2023.
\newblock Citation: A key to building responsible and accountable large language models.
\newblock \emph{arXiv preprint arXiv:2307.02185}.

\bibitem[{Huo et~al.(2023)Huo, Arabzadeh, and Clarke}]{huo2023retrieving}
Siqing Huo, Negar Arabzadeh, and Charles~LA Clarke. 2023.
\newblock Retrieving supporting evidence for llms generated answers.
\newblock \emph{arXiv preprint arXiv:2306.13781}.

\bibitem[{Izacard et~al.(2022)Izacard, Lewis, Lomeli, Hosseini, Petroni, Schick, Dwivedi-Yu, Joulin, Riedel, and Grave}]{izacard2022few}
Gautier Izacard, Patrick Lewis, Maria Lomeli, Lucas Hosseini, Fabio Petroni, Timo Schick, Jane Dwivedi-Yu, Armand Joulin, Sebastian Riedel, and Edouard Grave. 2022.
\newblock Few-shot learning with retrieval augmented language models.
\newblock \emph{arXiv preprint arXiv:2208.03299}.

\bibitem[{Ji et~al.(2023)Ji, Lee, Frieske, Yu, Su, Xu, Ishii, Bang, Madotto, and Fung}]{ji2023survey}
Ziwei Ji, Nayeon Lee, Rita Frieske, Tiezheng Yu, Dan Su, Yan Xu, Etsuko Ishii, Ye~Jin Bang, Andrea Madotto, and Pascale Fung. 2023.
\newblock Survey of hallucination in natural language generation.
\newblock \emph{ACM Computing Surveys}, 55(12):1--38.

\bibitem[{Kamalloo et~al.(2023)Kamalloo, Jafari, Zhang, Thakur, and Lin}]{kamalloo2023hagrid}
Ehsan Kamalloo, Aref Jafari, Xinyu Zhang, Nandan Thakur, and Jimmy Lin. 2023.
\newblock Hagrid: A human-llm collaborative dataset for generative information-seeking with attribution.
\newblock \emph{arXiv preprint arXiv:2307.16883}.

\bibitem[{Kamoi et~al.(2023)Kamoi, Goyal, Rodriguez, and Durrett}]{kamoi2023wice}
Ryo Kamoi, Tanya Goyal, Juan~Diego Rodriguez, and Greg Durrett. 2023.
\newblock Wice: Real-world entailment for claims in wikipedia.
\newblock \emph{arXiv preprint arXiv:2303.01432}.

\bibitem[{Kwiatkowski et~al.(2019)Kwiatkowski, Palomaki, Redfield, Collins, Parikh, Alberti, Epstein, Polosukhin, Devlin, Lee, Toutanova, Jones, Kelcey, Chang, Dai, Uszkoreit, Le, and Petrov}]{kwiatkowski-etal-2019-natural}
Tom Kwiatkowski, Jennimaria Palomaki, Olivia Redfield, Michael Collins, Ankur Parikh, Chris Alberti, Danielle Epstein, Illia Polosukhin, Jacob Devlin, Kenton Lee, Kristina Toutanova, Llion Jones, Matthew Kelcey, Ming-Wei Chang, Andrew~M. Dai, Jakob Uszkoreit, Quoc Le, and Slav Petrov. 2019.
\newblock \href {https://doi.org/10.1162/tacl_a_00276} {Natural questions: A benchmark for question answering research}.
\newblock \emph{Transactions of the Association for Computational Linguistics}, 7:452--466.

\bibitem[{Liu et~al.(2023)Liu, Zhang, and Liang}]{liu2023evaluating}
Nelson~F Liu, Tianyi Zhang, and Percy Liang. 2023.
\newblock Evaluating verifiability in generative search engines.
\newblock \emph{arXiv preprint arXiv:2304.09848}.

\bibitem[{Liu et~al.(2019)Liu, Ott, Goyal, Du, Joshi, Chen, Levy, Lewis, Zettlemoyer, and Stoyanov}]{liu2019roberta}
Yinhan Liu, Myle Ott, Naman Goyal, Jingfei Du, Mandar Joshi, Danqi Chen, Omer Levy, Mike Lewis, Luke Zettlemoyer, and Veselin Stoyanov. 2019.
\newblock Roberta: A robustly optimized bert pretraining approach.
\newblock \emph{arXiv preprint arXiv:1907.11692}.

\bibitem[{Malaviya et~al.(2023)Malaviya, Lee, Chen, Sieber, Yatskar, and Roth}]{malaviya2023expertqa}
Chaitanya Malaviya, Subin Lee, Sihao Chen, Elizabeth Sieber, Mark Yatskar, and Dan Roth. 2023.
\newblock Expertqa: Expert-curated questions and attributed answers.
\newblock \emph{arXiv preprint arXiv:2309.07852}.

\bibitem[{Metzler et~al.(2021)Metzler, Tay, Bahri, and Najork}]{metzler2021rethinking}
Donald Metzler, Yi~Tay, Dara Bahri, and Marc Najork. 2021.
\newblock Rethinking search: making domain experts out of dilettantes.
\newblock In \emph{Acm sigir forum}, volume~55, pages 1--27. ACM New York, NY, USA.

\bibitem[{Min et~al.(2023)Min, Krishna, Lyu, Lewis, Yih, Koh, Iyyer, Zettlemoyer, and Hajishirzi}]{min2023factscore}
Sewon Min, Kalpesh Krishna, Xinxi Lyu, Mike Lewis, Wen-tau Yih, Pang~Wei Koh, Mohit Iyyer, Luke Zettlemoyer, and Hannaneh Hajishirzi. 2023.
\newblock Factscore: Fine-grained atomic evaluation of factual precision in long form text generation.
\newblock \emph{arXiv preprint arXiv:2305.14251}.

\bibitem[{Nallapati et~al.(2016)Nallapati, Zhou, Gulcehre, Xiang et~al.}]{nallapati2016abstractive}
Ramesh Nallapati, Bowen Zhou, Caglar Gulcehre, Bing Xiang, et~al. 2016.
\newblock Abstractive text summarization using sequence-to-sequence rnns and beyond.
\newblock \emph{arXiv preprint arXiv:1602.06023}.

\bibitem[{Nenkova and Passonneau(2004)}]{nenkova-passonneau-2004-evaluating}
Ani Nenkova and Rebecca Passonneau. 2004.
\newblock \href {https://www.aclweb.org/anthology/N04-1019} {Evaluating content selection in summarization: The pyramid method}.
\newblock In \emph{Proceedings of the Human Language Technology Conference of the North {A}merican Chapter of the Association for Computational Linguistics: {HLT}-{NAACL} 2004}, pages 145--152, Boston, Massachusetts, USA. Association for Computational Linguistics.

\bibitem[{Nie et~al.(2019)Nie, Williams, Dinan, Bansal, Weston, and Kiela}]{nie2019adversarial}
Yixin Nie, Adina Williams, Emily Dinan, Mohit Bansal, Jason Weston, and Douwe Kiela. 2019.
\newblock Adversarial nli: A new benchmark for natural language understanding.
\newblock \emph{arXiv preprint arXiv:1910.14599}.

\bibitem[{OpenAI(2023)}]{chatgpt}
OpenAI. 2023.
\newblock {OpenAI}. (2023). {ChatGPT}.

\bibitem[{Raffel et~al.(2020)Raffel, Shazeer, Roberts, Lee, Narang, Matena, Zhou, Li, and Liu}]{raffel2020exploring}
Colin Raffel, Noam Shazeer, Adam Roberts, Katherine Lee, Sharan Narang, Michael Matena, Yanqi Zhou, Wei Li, and Peter~J Liu. 2020.
\newblock Exploring the limits of transfer learning with a unified text-to-text transformer.
\newblock \emph{The Journal of Machine Learning Research}, 21(1):5485--5551.

\bibitem[{Rajpurkar et~al.(2016)Rajpurkar, Zhang, Lopyrev, and Liang}]{rajpurkar2016squad}
Pranav Rajpurkar, Jian Zhang, Konstantin Lopyrev, and Percy Liang. 2016.
\newblock Squad: 100,000+ questions for machine comprehension of text.
\newblock \emph{arXiv preprint arXiv:1606.05250}.

\bibitem[{Rashkin et~al.(2023)Rashkin, Nikolaev, Lamm, Aroyo, Collins, Das, Petrov, Singh~Tomar, Turc, and Reitter}]{rashkin2023measuring}
Hannah Rashkin, Vitaly Nikolaev, Matthew Lamm, Lora Aroyo, Michael Collins, Dipanjan Das, Slav Petrov, Gaurav Singh~Tomar, Iulia Turc, and David Reitter. 2023.
\newblock Measuring attribution in natural language generation models.
\newblock \emph{Computational Linguistics}, pages 1--66.

\bibitem[{Shah and Bender(2022)}]{shah2022situating}
Chirag Shah and Emily~M Bender. 2022.
\newblock Situating search.
\newblock In \emph{Proceedings of the 2022 Conference on Human Information Interaction and Retrieval}, pages 221--232.

\bibitem[{Shapira et~al.(2019)Shapira, Gabay, Gao, Ronen, Pasunuru, Bansal, Amsterdamer, and Dagan}]{shapira-etal-2019-crowdsourcing}
Ori Shapira, David Gabay, Yang Gao, Hadar Ronen, Ramakanth Pasunuru, Mohit Bansal, Yael Amsterdamer, and Ido Dagan. 2019.
\newblock \href {https://doi.org/10.18653/v1/N19-1072} {Crowdsourcing lightweight pyramids for manual summary evaluation}.
\newblock In \emph{Proceedings of the 2019 Conference of the North {A}merican Chapter of the Association for Computational Linguistics: Human Language Technologies, Volume 1 (Long and Short Papers)}, pages 682--687, Minneapolis, Minnesota. Association for Computational Linguistics.

\bibitem[{Stanovsky et~al.(2018)Stanovsky, Michael, Zettlemoyer, and Dagan}]{stanovsky-etal-2018-supervised}
Gabriel Stanovsky, Julian Michael, Luke Zettlemoyer, and Ido Dagan. 2018.
\newblock \href {https://doi.org/10.18653/v1/N18-1081} {Supervised open information extraction}.
\newblock In \emph{Proceedings of the 2018 Conference of the North {A}merican Chapter of the Association for Computational Linguistics: Human Language Technologies, Volume 1 (Long Papers)}, pages 885--895, New Orleans, Louisiana. Association for Computational Linguistics.

\bibitem[{Tay et~al.(2022)Tay, Tran, Dehghani, Ni, Bahri, Mehta, Qin, Hui, Zhao, Gupta et~al.}]{tay2022transformer}
Yi~Tay, Vinh Tran, Mostafa Dehghani, Jianmo Ni, Dara Bahri, Harsh Mehta, Zhen Qin, Kai Hui, Zhe Zhao, Jai Gupta, et~al. 2022.
\newblock Transformer memory as a differentiable search index.
\newblock \emph{Advances in Neural Information Processing Systems}, 35:21831--21843.

\bibitem[{Wang et~al.(2019)Wang, Yu, Sun, Chen, Yu, McAllester, and Roth}]{wang2019evidence}
Hai Wang, Dian Yu, Kai Sun, Jianshu Chen, Dong Yu, David McAllester, and Dan Roth. 2019.
\newblock Evidence sentence extraction for machine reading comprehension.
\newblock \emph{arXiv preprint arXiv:1902.08852}.

\bibitem[{Weller et~al.(2023)Weller, Marone, Weir, Lawrie, Khashabi, and Van~Durme}]{weller2023according}
Orion Weller, Marc Marone, Nathaniel Weir, Dawn Lawrie, Daniel Khashabi, and Benjamin Van~Durme. 2023.
\newblock " according to..." prompting language models improves quoting from pre-training data.
\newblock \emph{arXiv preprint arXiv:2305.13252}.

\bibitem[{Yadav et~al.(2020)Yadav, Bethard, and Surdeanu}]{yadav-etal-2020-unsupervised}
Vikas Yadav, Steven Bethard, and Mihai Surdeanu. 2020.
\newblock \href {https://doi.org/10.18653/v1/2020.acl-main.414} {Unsupervised alignment-based iterative evidence retrieval for multi-hop question answering}.
\newblock In \emph{Proceedings of the 58th Annual Meeting of the Association for Computational Linguistics}, pages 4514--4525, Online. Association for Computational Linguistics.

\bibitem[{Yin et~al.(2021)Yin, Radev, and Xiong}]{yin-etal-2021-docnli}
Wenpeng Yin, Dragomir Radev, and Caiming Xiong. 2021.
\newblock \href {https://doi.org/10.18653/v1/2021.findings-acl.435} {{D}oc{NLI}: A large-scale dataset for document-level natural language inference}.
\newblock In \emph{Findings of the Association for Computational Linguistics: ACL-IJCNLP 2021}, pages 4913--4922, Online. Association for Computational Linguistics.

\bibitem[{Yue et~al.(2023)Yue, Wang, Zhang, Chen, Su, and Sun}]{yue2023automatic}
Xiang Yue, Boshi Wang, Kai Zhang, Ziru Chen, Yu~Su, and Huan Sun. 2023.
\newblock Automatic evaluation of attribution by large language models.
\newblock \emph{arXiv preprint arXiv:2305.06311}.

\bibitem[{Zhang et~al.(2022)Zhang, Thakur, Ogundepo, Kamalloo, Alfonso-Hermelo, Li, Liu, Rezagholizadeh, and Lin}]{zhang2022making}
Xinyu Zhang, Nandan Thakur, Odunayo Ogundepo, Ehsan Kamalloo, David Alfonso-Hermelo, Xiaoguang Li, Qun Liu, Mehdi Rezagholizadeh, and Jimmy Lin. 2022.
\newblock Making a miracl: Multilingual information retrieval across a continuum of languages.
\newblock \emph{arXiv preprint arXiv:2210.09984}.

\end{thebibliography}

\appendix
\clearpage
\section*{Appendix} \label{sec:appendix}

\section{Baseline Models} \label{sec:app-baselines}
\begin{itemize}
    \item{\textbf{BM25 (sparse)} is a classical bag-of-words based sparse retrieval method that relies on lexical overlap, term frequency heuristics, inverse document frequency and document length for retrieval relevant passages given a query.}
\item{\textbf{GTR (dense)} is a dense retrieval method that embeds both documents and queries into low-dimensional representations using T5-based~\citep{raffel2020exploring} dual encoders, with one of the encoders tailored to the queries and the other to the documents.}
\item{\textbf{MonoT5} is a T5-based model that takes a query and a document, and outputs the probability of relevance of document with respect to the query. The documents are ranked based on this probability.}
\end{itemize}

\begin{table*}[h]
\centering
\scriptsize
\begin{tabular}{c|p{7.2cm}|p{7.2cm}}
\toprule
\textbf{S.No.}&\textbf{Input} & \textbf{Output Decompositions}\\
\midrule
 \multicolumn{3}{c}{\multirow{1}{*}{\textbf{Correct Decompositions}}} \\
\midrule
1. & Based on the given contexts, the University Library of the Library of Congress in Washington D.C. is the biggest library in the world, with more than 167 million items including books, printed materials, recordings, photographs, maps, sheet music, and manuscripts. & The University Library of the Library of Congress is in Washington D.C. The University Library of the Library of Congress is the biggest library in the world. The University Library of the Library of Congress has more than 167 million items. The items include books, printed materials, recordings, photographs, maps, sheet music, and manuscripts.\\
2. & The Islamic Golden Age is traditionally dated from the 8th century to the 14th century. & The Islamic Golden Age is traditionally dated. The Islamic Golden Age is traditionally dated from the 8th century. The Islamic Golden Age is traditionally dated to the 14th century.\\
\midrule
 \multicolumn{3}{c}{\multirow{1}{*}{\textbf{Erroneous Decompositions}}} \\
\midrule
1. & Ronnie Van Zant was born in Jacksonville, Florida.& Ronnie Van Zant was born. Ronnie Van Zant was born in Jacksonville, Florida\\
2. & The Beer Store, also known as Brewers Retail Inc, was founded in 1927. & The Beer Store is also known as Brewers Retail Inc. The Beer Store was founded in 1927. The Beer Store was founded in Ontario, Canada\\
\bottomrule
\end{tabular}
\caption{Sample outputs from the Answer Decomposer. 1. shows over-decomposition, and 2. shows hallucination error under Erroneous Decompositions.} \label{tab:decompose-examples}
\end{table*}

\section{Entailment model DocNLI} \label{sec:app-docnli}
We have used RoBERTa-L model trained on DocNLI dataset as our go-to entailment model. DocNLI contains an array of reformulated versions of existing datasets (adversarial NLI (ANLI)~\citep{nie2019adversarial}, the question answering benchmark SQuAD~\cite{rajpurkar2016squad} and three summarization benchmarks (CNN/DailyMail~\citep{nallapati2016abstractive}, DUC2001\footnote{\url{https://www-nlpir.nist.gov/projects/duc/guidelines/2001.html}}, and Curation~\citep{curation2020})) by transforming various summarization and question answering datasets into natural language inference form to ensure that the premise and hypothesis are paragraph-level long and that the dataset does not contain any artifacts such as hypothesis length bias, direct overlap between premise and hypothesis. Table~\ref{tab:docnli-performance} reports the accuracy results of DocNLI on various NLI datasets.
\begin{table*}[h]
\centering
\scriptsize
\begin{tabular}{ccccc}
\toprule
\textbf{FEVER}&	\textbf{MCTest (v160)}&	\textbf{MCTest (v500)}	& \textbf{SciTail}	& \textbf{MNLI}\\
\midrule
$88.84$&	$90.00$	&$85.83$	&$78.17$	&$91.13$\\
\bottomrule
\end{tabular}
\caption{Accuracy of DocNLI (used as the Attributor in our work) model on various NLI datasets. We report the numbers as-is from~\citet{yin-etal-2021-docnli}.} \label{tab:docnli-performance}
\end{table*}
\section{Additional Results}
We present results on verifiability dataset when top 150 sentences retrieved using GTR and MonoT5 are used by \texttt{ADiOSAA} and when all the sentences in a document are used in Table~\ref{tab:add-results-verifiability}. In all the cases, \texttt{ADiOSAA} outperforms all the ablations - removing decompose, optimal selection or both. This shows that each of the components of the proposed approach is important for the attribution task.
\begin{table*}[h]
\centering
\resizebox{0.99\linewidth}{!}{
\begin{tabular}{l|ccc|ccc|ccc}
\toprule
\multirow{2}{*}{\centering{\textbf{Model}}} & \multicolumn{3}{c|}{\centering{\textbf{Top 1}}} & \multicolumn{3}{c|}{\centering{\textbf{Top 2}}} & \multicolumn{3}{c}{\centering{\textbf{Top 4}}}\\
\cline{2-10}\\
& \textbf{P}& \textbf{R} & \textbf{F1} & \textbf{P}& \textbf{R} & \textbf{F1} & \textbf{P}& \textbf{R} & \textbf{F1}\\
\midrule
All + \texttt{ADIOSAA}&$\mathbf{0.537}$&$\mathbf{0.422}$&$\mathbf{0.452}$&$\mathbf{0.479}$&$\mathbf{0.540}$&$\mathbf{0.482}$&$\mathbf{0.471}$&$\mathbf{0.598}$&$\mathbf{0.494}$\\
All + \texttt{ADIOSAA} - Decomposer&$0.462$&$0.381$&$0.404$&$0.435$&$0.408$&$0.402$&$0.433$&$0.408$&$0.401$\\
All + \texttt{ADIOSAA} - Optimal Selection&$0.368$&$0.289$&$0.311$&$0.272$&$0.327$&$0.279$&$0.250$&$0.353$&$0.270$\\
All + \texttt{ADIOSAA} - Decomposer - Optimal Selection&$0.262$&$0.226$&$0.236$&$0.262$&$0.226$&$0.236$&$0.262$&$0.226$&$0.236$\\
\midrule
GTR + \texttt{ADIOSAA}&$\mathbf{0.538}$&$\mathbf{0.423}$&$\mathbf{0.453}$&$\mathbf{0.479}$&$\mathbf{0.541}$&$\mathbf{0.483}$&$\mathbf{0.471}$&$\mathbf{0.598}$&$\mathbf{0.494}$\\
GTR + \texttt{ADIOSAA} - Decomposer&$0.463$&$0.382$&$0.405$&$0.435$&$0.409$&$0.403$&$0.433$&$0.409$&$0.402$\\
GTR + \texttt{ADIOSAA} - Optimal Selection&$0.372$&$0.294$&$0.315$&$0.275$&$0.332$&$0.282$&$0.252$&$0.358$&$0.273$\\
GTR + \texttt{ADIOSAA} - Decomposer - Optimal Selection&$0.265$&$0.229$&$0.238$&$0.265$&$0.229$&$0.238$&$0.265$&$0.229$&$0.238$\\
\midrule
MonoT5 + \texttt{ADIOSAA}&$\mathbf{0.537}$&$\mathbf{0.422}$&$\mathbf{0.452}$&$\mathbf{0.479}$&$\mathbf{0.540}$&$\mathbf{0.482}$&$\mathbf{0.471}$&$\mathbf{0.598}$&$\mathbf{0.494}$\\
MonoT5 + \texttt{ADIOSAA} - Decomposer&$0.467$&$0.385$&$0.408$&$0.439$&$0.412$&$0.407$&$0.437$&$0.413$&$0.406$\\
MonoT5 + \texttt{ADIOSAA} - Optimal Selection	&$0.371$&$0.292$&$0.314$&$0.274$&$0.330$&$0.281$&$0.251$&$0.356$&$0.272$\\
MonoT5 + \texttt{ADIOSAA} - Decomposer - Optimal Selection&$0.265$&$0.229$&$0.238$&$0.265$&$0.229$&$0.238$&$0.265$&$0.229$&$0.238$\\
\bottomrule
\end{tabular}
}
\caption{Evaluation results with GTR, MonoT5 and all sentences for Verifiability dataset.} \label{tab:add-results-verifiability}
\end{table*}
\subsection{Implementation Details}
The RoBERTa-L model contains 355 million parameters. We use off-the-shelf model so no training is required. We only perform inference on one NVIDIA T4 16GB GPU machine. All the reported scores are from one run of the model inference. Hyperparameter tuning is done for $\delta$ and entailment probability threshold using the development set of Verifiability dataset.

\end{document}